\documentclass{article}
\usepackage{amsmath,epsfig}
\usepackage[preprint]{spconfa4}
\usepackage{algorithm}
\usepackage{algorithmic}
\usepackage{hyperref}
\usepackage[hyphenbreaks]{breakurl}
\usepackage[misc]{ifsym}

\copyrightnotice{978-1-6654-3864-3/21/\$31.00 ©2021 IEEE}

\let\OLDthebibliography\thebibliography
\renewcommand\thebibliography[1]{
  \OLDthebibliography{#1}
  \setlength{\parskip}{0pt}
  \setlength{\itemsep}{0pt plus 0.3ex}
}

\begin{document}\sloppy

\def\x{{\mathbf x}}
\def\L{{\cal L}}
\renewcommand{\algorithmicrequire}{ \textbf{Input:}} 
\renewcommand{\algorithmicensure}{ \textbf{Output:}} 

\title{RPAttack: Refined Patch Attack on General Object Detectors}
%
\name{Hao Huang$^{\ast}$, Yongtao Wang$^{\ast}$\textsuperscript{\Letter}, Zhaoyu Chen$^{\dagger}$, Zhi Tang$^{\ast}$, Wenqiang Zhang$^{\dagger}$, Kai-Kuang Ma$^{\ddagger}$}
\address{$^{\ast}$Peking University, Beijing, China; $^{\dagger}$Academy for Engineering and Technology, Fudan University, \\Shanghai, China; and $^{\ddagger}$Nanyang Technological University, Singapore.
\\
huanghao@stu.pku.edu.cn; wyt@pku.edu.cn; zhaoyuchen20@fudan.edu.cn; \\ 
tangzhi@pku.edu.cn; wqzhang@fudan.edu.cn; ekkma@ntu.edu.sg}

\maketitle

\begin{abstract}
Nowadays, general object detectors like YOLO and Faster R-CNN as well as their variants are widely exploited in many applications. Many works have revealed that these detectors are extremely vulnerable to adversarial patch attacks. The perturbed regions generated by previous patch-based attack works on object detectors are very large which are not necessary for attacking and perceptible for human eyes. To generate much less but more efficient perturbation, we propose a novel patch-based method for attacking general object detectors. Firstly, we propose a patch selection and refining scheme to find the pixels which have the greatest importance for attack and remove the inconsequential perturbations gradually.
Then, for a stable ensemble attack, we balance the gradients of detectors to avoid over-optimizing one of them during the training phase. Our RPAttack can achieve an amazing missed detection rate of 100\% for both Yolo v4 and Faster R-CNN while only modifies 0.32\% pixels on VOC 2007 test set. Our code is available at \url{https://github.com/VDIGPKU/RPAttack}.
\end{abstract}

\let\thefootnote\relax\footnotetext{This work was supported by National Key R\&D Program of China No. 2019YFB1406302. This work was also a research achievement of Key Laboratory of Science, Technology and Standard in Press Industry (Key Laboratory of Intelligent Press Media Technology).}

\section{Introduction}
\label{sec:intro}

Object detection is a fundamental computer vision task which needs to predict the category and location of the object simultaneously. With the great development of deep learning, object detectors are successfully integrated into more and more real-world application systems. Hence, ensuring the safe usage of object detectors becomes a very important problem to be tackled. 

Many works\cite{XieWZZXY17,ZhangZL20} have revealed the vulnerability of general object detectors by generating adversarial perturbations on the whole image. Though these perturbations are invisible for human eyes, they can not be performed in real-world cases, since it is impossible to attack the whole scene shown in the image. On the other hand, patch-based attack methods\cite{abs-1712-09665, LiuYLSCL19, Object, DPAttack} could be exploited for real-world attacks that only modify some patches rather than the whole image. 
However, the adversarial patches generated by these methods are so large thus are noticeable for human eyes. Besides, some patch-based methods like DPATCH\cite{LiuYLSCL19} is not efficient enough which needs to train 200k iterations to generate an adversarial patch.

\begin{figure}[t]
\begin{minipage}[b]{1.0\linewidth}
  \centering
  \includegraphics[width=8cm]{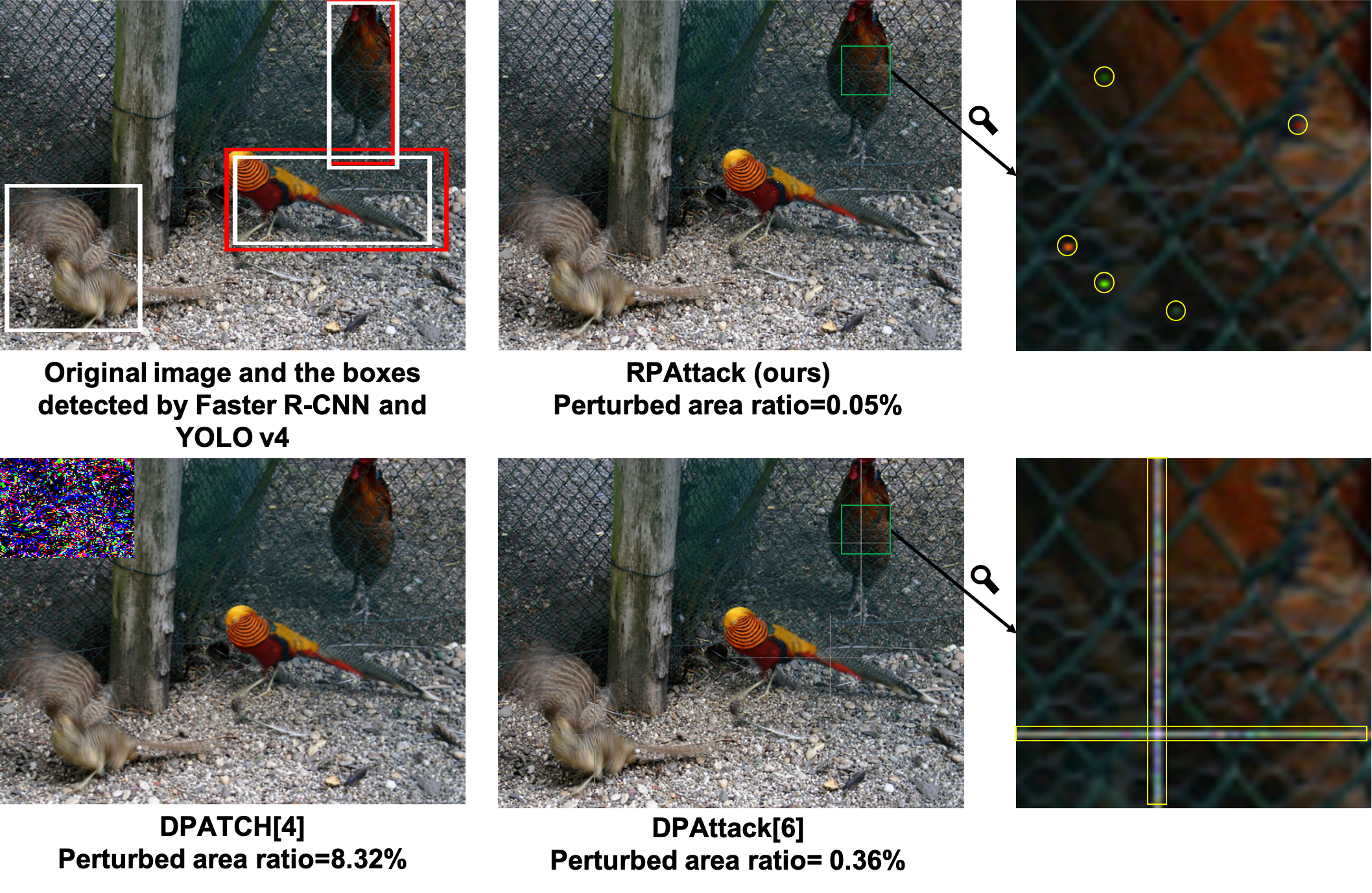}
  \caption{Illustration of an example. Boxes with red lines and white lines are predicted by YOLO v4\cite{abs-2004-10934} and Faster R-CNN \cite{RenHG017} respectively. We show adversarial images generated by RPAttack (ours), DPATCH and DPAttack. Noticeably, the proposed RPAttack has generated the least perturbations. We also show some local perturbation clearly for comparison with DPAttack. Moreover, no object can be detected from the adversarial image generated by our RPAttack, for both YOLO v4 and Faster R-CNN.}
  \label{fig:demo}
\end{minipage}
\end{figure}

Obviously, for an attack, it is better that less area of the image are perturbed while the attack effect doesn’t degenerate. Moreover, we empirically observe that the importance of different pixels in an image or a patch varies greatly for adversarial attacking. To this light, we try to find the pixels which have the greatest importance for the attack, which we call \textbf{key-pixels}. We propose a patch selection and refining scheme to gradually achieve this goal since we learn that at the beginning of the training process, the key-pixels are extremely difficult to be identified and would be changed when we update the perturbations. Specifically, we adaptively find the patches for attack based on the gradients and gradually remove the inconsequential pixels in each patch until the attack process gets stable. Experiments show that our proposed method is very effective and efficient, that is, it can decrease the detection accuracy (i.e., mAP) of both YOLO v4 and Faster R-CNN to 0 within only 2k training iterations. Besides, most recent works\cite{LiuYLSCL19, ZhangZL20} can only attack a specific detector while our proposed ensemble RPAttack can attack YOLO v4\cite{abs-2004-10934} and Faster R-CNN\cite{RenHG017}, i.e., two detectors with different architectures, at the same time. Specifically, we balance the gradients of both detectors to avoid over-optimizing one of them during the training phase. Figure 1 illustrates one example, and one can observe that the perturbations for this example generated by our proposed RPAttack are negligible while successfully fool both two detectors, that is, no object has been detected by them. 
To sum up, the contributions of this work are threefold:
\begin{itemize}
  \item We propose a novel method for attacking general object detectors and achieve an amazing missed detection rate of 100\% for both Yolo v4 and Faster R-CNN while only modify 0.32\% pixels on VOC 2007 test set.
  
   \item We first propose a patch selection and refining scheme for finding key-pixels dynamically and gradually remove the inconsequential perturbations. 
 
  \item We further propose ensemble attacks on YOLO v4 and Faster R-CNN simultaneously by balancing the gradients of both detectors to avoid over-optimizing one of them during the training phase.
\end{itemize}

\section{Related Work}
\subsection{General Object Detection}
In recent years, there has been great progress in the field of general object detection. Deep learning methods\cite{abs-2004-10934,RenHG017} have greatly improved the performance in object detection. The mainstream methods based on deep learning can be roughly divided into one-stage methods like YOLO\cite{YOLOV1} and two-stage methods like Faster R-CNN\cite{RenHG017}.

In this paper, we attack two detectors, i.e., YOLO v4, and Faster R-CNN, which are the most representative ones of one-stage detectors and two-stage detectors respectively. Specifically, YOLO v4 regresses bounding boxes and predicts the class probabilities directly after a single pass of input, while Faster R-CNN first produces proposals with a region proposal network (RPN) and then classifies and re-regresses these proposals with a detection head.

\subsection{Adversarial Examples and Patches}

 The adversarial examples are first proposed in \cite{SzegedyZSBEGF13}, revealing the vulnerability of classification neural networks. Adversarial examples of networks are the input data with deliberate perturbations. Although the perturbations are too small to be noticeable by human eyes, it can successfully mislead various deep learning-based models. \cite{abs-1712-09665} first advances the adversarial patches, which can also fool the classification networks. The previous works on adversarial patches mainly concentrate on classification tasks, thus not suitable for the object detection task which needs to predict the category and location of the object instances simultaneously. DPATCH\cite{LiuYLSCL19} proposes adversarial patches to disable object detectors, which can decrease the mAP greatly on YOLO\cite{YOLOV1} and Faster R-CNN. However, the adversarial patches produced by it are usually very large in size, which are inefficient and noticeable for human eyes. DPAttack\cite{DPAttack} designs a diffused patch of asteroid-shaped or grid-shaped based on the detected results and pays more attention to unsuccessfully attacked proposals. Object Hider\cite{Object} uses a heatmap-based and consensus-based algorithm to select patches for the attack. Compared with these two relevant works, our RPAttack can generate patches with much less perturbation while achieving better attack performance. 

\begin{figure*}[t]
\begin{minipage}[b]{1.0\linewidth}
  \centering
  \includegraphics[width=16cm]{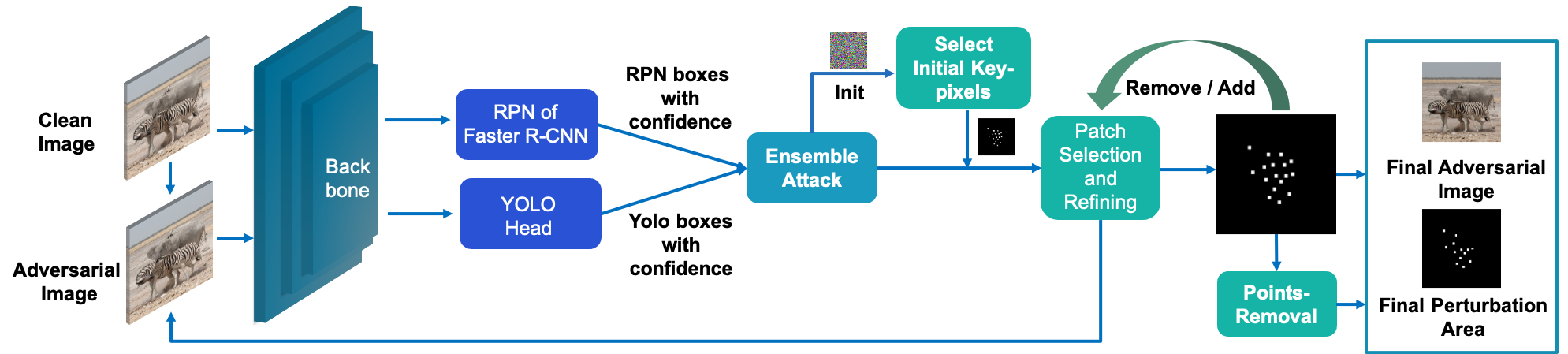}
  \caption{The pipeline of the proposd RPAttack.}\medskip
  \label{fig:pipeline}
\end{minipage}
\end{figure*}

\section{Method}
In this section, we introduce our RPAttack in detail. Firstly, we give the problem formulation in Section 3.1. After that, the process of RPAttack is presented in Section 3.2. Then, we describe a novel patch selection and refining scheme in Section 3.3. Finally, in Section3.4, we discuss how to balance the gradients from different detectors to stabilize the ensemble attack.

\subsection{Problem Formulation}
In this work, we attack two detectors of different architectures, Faster R-CNN and YOLO v4, and try to hide all the objects from these two detectors. One-stage detectors like YOLO v4 directly regress the bounding boxes with the confidence scores, and two-stage detectors like Faster R-CNN use RPN to get bounding boxes of proposals with the confidence scores. We use ${b_i, c_i}$ to denote the bounding box and the confidence score. Our goal is to hide all the objects from these two detectors with the least perturbation, which can be formulated as
\begin{eqnarray}
\min_{P^j}  \sum_{j}^{m} \sum_{i}^{n} D_i(x_j+{{P^j}}) + \sum_{j}^{m} Area(P^j),
\end{eqnarray}
where $x_j$ is the $j$th image, $D_i(x)$ represents the bounding box number of image $x$ detected by $i$th detector, ${P^j}$ is the perturbation we add to the $j$th image, and $Area(P^j)$ is the area of perturbation in $j$th image.

\subsection{Attack on General Object Detectors}
In this section, we introduce the detailed attack process of our RPAttack.
To hide the objects from different detectors, we need to reduce the confidence score $c_i$ of each bounding box. Based on this, we define the loss function as:
\begin{eqnarray}
    J(c)=- \frac{1}{k}\cdot \sum_i^k L(c_{i}, 0),
\end{eqnarray}
where $L(\cdot,\cdot)$ is the Mean Square Error (MSE), $c_i$ is the confidence score of $i$th bounding box.

\begin{figure}[t]
\begin{minipage}[b]{1.0\linewidth}
  \centering
  \includegraphics[width=7.8cm]{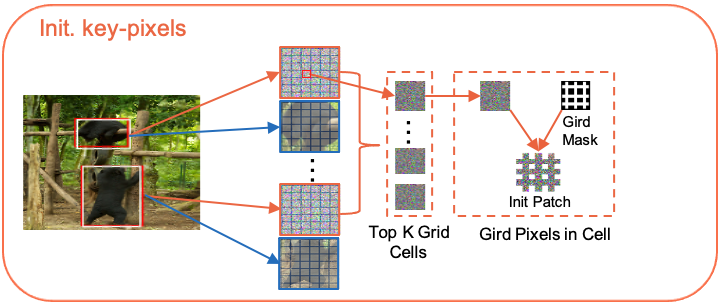}
  \caption{Illustration of the process to generate initial key-pixels.}\medskip
  \label{fig:init}
\end{minipage}
\end{figure}

We attack Faster R-CNN and YOLO v4 at the same time, and the whole attack pipeline is shown in the Figure \ref{fig:pipeline}. Firstly, we input the image to two detectors and get the bounding boxes with confidence scores. Then, we perform an instance-level attack which means we can modify all pixels in the bounding boxes. The gradient at one iteration does not reveal the location of key-pixels accurately so we repeat instance-level attack until all detectors can not detect any object for a stable and credible gradient heat map to find initial key-pixels. Next, we attack the original image with the initial key-pixels obtained from the cumulative gradient heat map. Then, we perform an ensemble attack in which we adaptively find the most suitable location to modify in every $A_k$ iteration and remove the inconsequential perturbations when the attack gets stable. Finally, for generating less perturbed pixels, we perform points-removal to further remove the perturbed pixels which have no effect on the final results. After the above attack process, we can get an adversarial image with much less perturbation and better performance.

\begin{figure}[t]
\begin{minipage}[b]{1.0\linewidth}
  \centering
  \includegraphics[width=9cm]{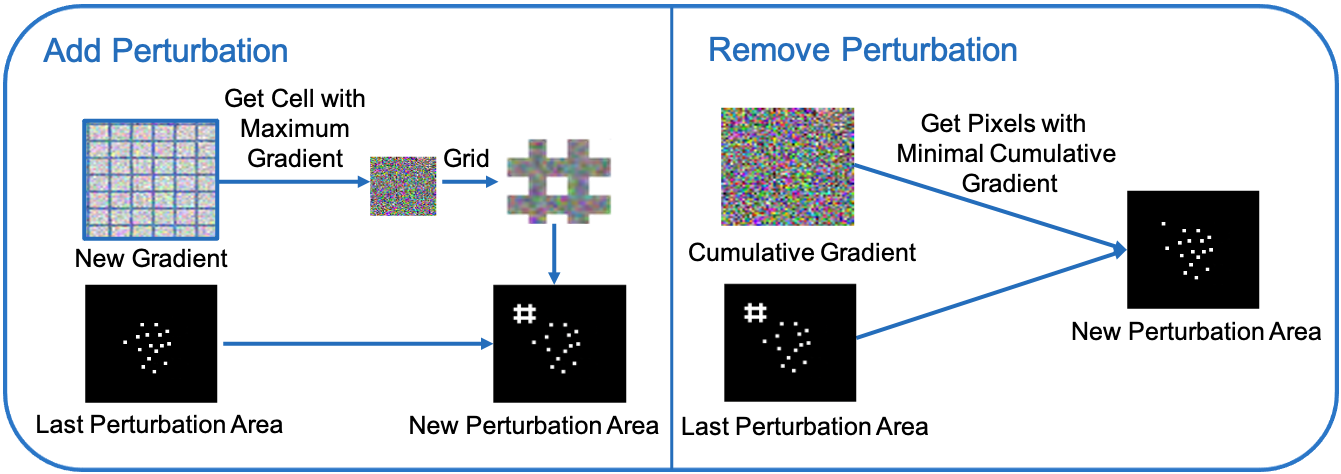}
  \caption{Illustration of the proposed patch selection and refining scheme.}\medskip
  \label{fig:DPSM}
\end{minipage}
\end{figure}

The initial key-pixels are determined by the cumulative gradient, as shown in Figure \ref{fig:init}. Specifically, we divide each $H \times W$ instance box predicted from the original image into an $\lceil  H/m  \rceil \times \lceil W/m \rceil$ grid of cells(the size of each is $m \times m$). Then, we sum the absolute gradient values in every cell of all instance boxes and select top $K$(we set $K$=5 in our experiments) cells. As the adversarial patches of grid shape can successfully attack an image with the least perturbed pixels area in our experiments, we transfer the top $K$ cells to grid shaped ones and regard the pixels on the grid lines as initial key-pixels.

Our RPAttack is based on the Iterative Fast Gradient Sign Method (I-FGSM)\cite{KurakinGB17a}, iteratively applies FGSM\cite{GoodfellowSS14} with a small step size $\alpha$. Moreover, to stabilize the ensemble attack process, we introduce extra parameters that balance gradients of different detectors to I-FGSM, and the details are presented in Section 3.4.

\subsection{Patch Selection and Refining}

To find key-pixels for attack and remove the inconsequential perturbations, we propose a novel patch selection and refining scheme as Figure \ref{fig:DPSM} shows.

Though we can get the initial location of key-pixels based on the cumulative gradient heat map, the key-pixels will be changed when we update the perturbations. Besides, the initial perturbations may not include all key-pixels. An adaptive method is proposed to solve these problems, that is, we add an adversarial patch to the image based on the current gradient in every $A_k$ iterations. In this way, we can find the new key-pixels which are the most suitable ones in the current iteration. 

The pixels we modified are not all playing a key role in an attack. However, removing some of them after the attack process may cause a decrease in attack performance because other key-pixels haven't been modified sufficiently. Removing the inconsequential perturbation during the attack process and keeping the attack process, can modify the remaining key-pixels sufficiently. In our work, when the attack gets stable (the number of bounding boxes reaches the minimum at least $D_k$ times), we remove the perturbations where the value of pixels changed is less than the average changed value divided by 3. We repeat the process until reaching the maximum number of iterations $I$. 


\begin{algorithm}[htb]
\caption{Patch Selection and Refining Scheme} 
\label{alg1}
\begin{algorithmic}[1]

\REQUIRE ~~
 $A_k$(the frequency of adding perturbation), $D_k$(the threshold of decreasing perturbation), $I$(maximum iterations), $P$(adversarial perturbation), $x_k^*$(an adversarial image); $PN$(the number of patches), $Attack$(attack method which returns new perturbations and the number of bounding boxes), $MP$(max number of patches).

\ENSURE ~~
adversarial perturbation $P$

\STATE $state\_add \Leftarrow false, state\_decrease \Leftarrow false$
\STATE $a_k, \Leftarrow 0,  d_k \Leftarrow 0, min\_bb\_num  \Leftarrow INF$

\FOR{each $i \in [1,I]$}
\STATE $P, N \Leftarrow Attack(P, x_k^*)$
\IF{$N > 0$}
\STATE $a\_k \Leftarrow a\_k + 1$
\IF{$a_k \% A_k = 0 \ and \ PN < MP$}
\STATE $P \Leftarrow add \ patch(P), PN \Leftarrow PN+1$
\ENDIF
\ENDIF

\IF{$N < min\_bb\_num$}
\STATE $min\_bb\_num \Leftarrow N$
\ENDIF

\IF{$N = min\_bb\_num$}
\STATE $d_k \Leftarrow d_k + 1$
\IF{$d_k\%D_k=0$}
\STATE $state\_decrease \Leftarrow true$
\ENDIF
\IF{$state\_decrease$}
\STATE $P \Leftarrow decrease \ perturbation(P)$
\ENDIF
\ENDIF
\ENDFOR
\end{algorithmic}
\end{algorithm}

The details of the scheme can be referred to Algorithm \ref{alg1}. With the help of this scheme, the key-pixels can be found adaptively, and inconsequential perturbation can be removed. It not only greatly improves the performance of our RPAttack but also generates much less perturbation than \cite{LiuYLSCL19,Object,DPAttack}. 

\subsection{Stabilize Ensemble Attack}

Most recent works can only attack a specific detector while our RPAttack aims to attack YOLO v4 and Faster R-CNN simultaneously. However, the gradient produced by each detector will affect the other. To stabilize the ensemble attack and avoid over-optimizing a specific detector, we use the following method to modify images,
\begin{eqnarray}
    & x^*_{k+1}=x^*_{k}+ \sum_{i=1}^N w_i \cdot \alpha \cdot \mathrm{sign}(\nabla_xJ_i(x^*_k,y))) , \\
    & w_i = max(1, D_i(x^*) - D_i(x)),
\end{eqnarray}
where $x^*_{k}$ is adversarial example in $k$th iteration, $J_i$ is the loss function we use to attack $i$th detector, $\alpha$ is the updated step, the weight $w_i$ is the parameter of balancing gradients and $D_i(x)$ is the number of instances in $x$ detected by $i$th detector.

\section{Experiment}

In this section, we first describe the datasets as well as our implementation details. Then we present experimental results of our RPAttack. Finally, we conduct ablation studies to demonstrate the effectiveness of our methods in detail.

\begin{table}[]
\scriptsize
\centering
\caption{The RPAttack Results on VOC2007. $Y \overline{BS}$ and $F \overline{BS}$ denote the $\overline{BS}$ of attacking YOLO v4 and Faster R-CNN respectively..} 
\label{VOC2007}
\begin{tabular}{llllll}
\hline
Methods &      $\overline{AS}$ & Y $\overline{BS}$      & F $\overline{BS}$ & $\overline{OS}$ & mAP    \\ \hline
RPAttack   &    1.776& 1.000             & 1.000            & 3.553  & 0.00  \\
RPAttack+points-removal & 1.839& 1.000             & 1.000            & 3.679  & 0.00 \\ \hline
\end{tabular}
\end{table}

\begin{table}[]
\scriptsize
\centering
\caption{Comparison on Ali-Attack-Data, RPAttack($<=$10) means we limit the number of perturbed regions to no more than 10. $Y \overline{OS}$ and $F \overline{OS}$ denote the $\overline{OS}$ of attacking YOLO v4 and Faster R-CNN respectively.} 
\label{alidata}
\begin{tabular}{llll}
\hline
Methods          & Y $\overline{OS}$      & F $\overline{OS}$ & $\overline{OS}$    \\ \hline
Object Hider\cite{Object}     & -             & -             & 2.760   \\
DPAttack\cite{DPAttack}         & 1.563          & 1.436          & 2.999          \\
RPAttack($<=$10)  & 1.615          & 1.512          & 3.127          \\ 
RPAttack($<=$10)+points-removal & \textbf{1.627} & \textbf{1.522} & \textbf{3.149} \\
RPAttack & 1.723          & 1.721          & 3.444          \\
RPAttack+points-removal & \textbf{1.789} & \textbf{1.784} & \textbf{3.573} \\ \hline
\end{tabular}
\end{table}

\begin{table*}[]
\scriptsize
\centering
\caption{Detailed comparison of CAP, DPATCH, and RPAttack on VOC 2007 test set. All methods attack Faster R-CNN with ResNet101.} 
\label{com}
\scalebox{0.85}{
\begin{tabular}{lllllllllllllllllllll}
\hline
Methods   & aero & bike & bird & boat & bottle & bus & car & cat & chair & cow & table & dog & horse & mbike & person & plant & sheep & sofa & train & tv \\ \hline
No Patch & 88.80  & 84.60  & 84.80  & 71.00  & 80.40   & 88.70 & 83.20 & 87.20 & 67.60   & 87.80 & 67.00   & 85.00 & 89.00  & 87.40   & 87.00    & 57.00   & 87.40   & 79.00  & 85.50   & 84.70 \\
CAP\cite{ZhangZL20} & \textbf{0.00}  & 9.10  & \textbf{0.00}  & \textbf{0.00}  & \textbf{0.00}    & \textbf{0.00} & 1.50 & \textbf{0.00} & \textbf{0.00}   & \textbf{0.00} & \textbf{0.00}   & \textbf{0.00} & \textbf{0.00}   & 3.00   & 9.10    & \textbf{0.00}   & \textbf{0.00}   & \textbf{0.00}  & 9.10   & \textbf{0.00} \\
DPATCH\cite{LiuYLSCL19}  & 0.02 & \textbf{0.00}  & \textbf{0.00}  & \textbf{0.00}  & \textbf{0.00}    & 0.53 & 0.08 & 0.61 & \textbf{0.00}  & 0.02 & \textbf{0.00}  & 9.09 & 0.16  & \textbf{0.00}  & 9.09   & 0.16  & \textbf{0.00}  & 9.09 & \textbf{0.00}  & \textbf{0.00} \\
RPAttack(ours)  & \textbf{0.00}  & \textbf{0.00}  & \textbf{0.00}  & \textbf{0.00}  & \textbf{0.00}    & \textbf{0.00} & \textbf{0.00} & \textbf{0.00} & \textbf{0.00}   & \textbf{0.00} & \textbf{0.00}   & \textbf{0.00} & \textbf{0.00}   & \textbf{0.00}   & \textbf{0.00}    & \textbf{0.00}   & \textbf{0.00}   & \textbf{0.00}  & \textbf{0.00}   & \textbf{0.00} \\
\hline
\end{tabular}}
\end{table*}

\subsection{Datasets and Implementation Details}

We use two datasets in our experiments: VOC2007\cite{voc} test set, and the dataset in Alibaba-Tsinghua Adversarial Challenge on Object Detection which sample 1000 images from MS COCO 2017 test set\cite{coco}. We call the latter Ali-Attack-Data for simplicity.
As for the parameter setting of our method, we set the maximum iteration number $I=2k$ for VOC2007 and $I=4k$ for Ali-Attack-Data, the frequency of adding perturbation $A_k=100$, the threshold of decreasing perturbation $D_k=25$, the size of the patch is 70*70, the $P_{limit}$ is 0.02 and the backbone of Faster R-CNN is ResNet101.

In order to clearly show the performance of our algorithm, we introduce the following metrics: 
\begin{eqnarray}
& AS = 
\begin{cases}
\sum_{j}^{m} (2 -  \frac{{P^j}_{rate}} {P_{limit}}),& {P^j}_{rate} \leq P_{limit},\\
0,& \text{otherwise,} \\
\end{cases} \\
& BS = \sum_{j}^m \sum_{i}^{n} max(D_i(x_j) - D_i(x_j + {P^j}), 0), & \\
& OS = \sum_{j}^m \sum_{i}^{n} AS_j \cdot BS_{ij}, &
\end{eqnarray}
where $P_{limit}$ is the upper bound of the perturbation rate, ${P^j}_{rate}$ is the perturbation rate in $j$th adversarial image. AS shows the area score of perturbation. Specially, if the ${P^j}_{rate}>P_{limit}$, $AS_j=0$. BS is the difference between the number of bounding boxes predicted in original images and in adversarial images. OS shows the overall performance. Obviously, to achieve a high socre of $OS$, we need to add less perturbation while decrease more bounding boxes. Due to the different size of datasets, we use $\overline{AS}$, $\overline{BS}$ and $\overline{OS}$ to represent the average score per image in our results. Besides, to compare with some previous works, the decrease of $mAP$ is also used as an evaluation metric.

\subsection{The Results of RPAttack}

We show the results of RPAttack on VOC 2007 test set in Table \ref{VOC2007}. Our RPAttack can hide all objects from YOLO v4 and Faster R-CNN but only modify 0.45\%(average) pixels on VOC 2007 test set. After points-removal, the rate of modified pixels can drop to 0.32\%. These results reveal that modifying very few pixels can disable the SOTA detectors completely. Obviously, all objects are hidden so mAP drops from 81.7 to 0.

We noticed that some recent patch-based works\cite{Object, DPAttack} are also dedicated to decreasing the area of perturbation. They use clever strategies and achieve good performance in Ali-Attack-Data. These two works have the same task as ours but limit the connected regions of perturbed pixels to no more than 10. To compare with these works fairly, we follow their experimental settings and also limit the number of connected perturbed regions generated by our RPAttack to no more than 10. Despite adding such constraint, the experimental results show that the proposed RPAttack performs better and generates less perturbation compared with \cite{Object, DPAttack}, as shown in Table \ref{alidata}. Moreover, if we remove this constraint, we can achieve even more exciting results that we successfully hide 99.9\% objects from YOLO v4 and 99.6\% objects from Faster R-CNN with only 0.42\% perturbation.


DPATCH\cite{LiuYLSCL19} and CAP\cite{ZhangZL20} are representative patch-based and full-image-based attacks, respectively. The task of these methods is minimizing mAP by add patch-based or full-image-based perturbation. Any change in category or bounding box will cause the mAP to drop, making the task very simple compared to ours(hiding all objects). Hiding the objects will lead the detectors outputs nothing, which can decrease mAP though it is not our major goal. We are excited to discover that compared to DPATCH\cite{LiuYLSCL19} and CAP\cite{ZhangZL20}, we can decrease mAP even more as Table \ref{com} shows. 


\subsection{Ablation Study}

\begin{table}[]
\scriptsize
\centering
\caption{Ablation Study on VOC2007 test set (\textbf{unconstrained})} 
\label{ablation1}
\begin{tabular}{lllllll}
\hline
Random Location      &$\surd$& &   &      &      &    \\
Center BBOX Location     & &$\surd$&    &      &      &    \\
Gradient-based Location&  &  &  $\surd$&$\surd$& $\surd$& $\surd$     \\
Patch Selection and Refining   &  &      &  &$\surd$&     & $\surd$    \\
Stabilizing gradient   &  &    &      & &$\surd$    & $\surd$    \\ \hline \\
Y $\overline{BS}$       &0.985&0.980& 0.943 & \textbf{1.000} & 0.952 & \textbf{1.000} \\
F $\overline{BS}$      &0.877&0.729 & 0.892 & \textbf{1.000} & 0.896 & \textbf{1.000} \\
$\overline{AS}$      &1.000   &1.000 &1.000 & 1.834 & 1.000 & \textbf{1.835} \\
$\overline{OS}$            &1.862 &1.709& 1.835 & 3.670 & 1.848 & \textbf{3.671} \\ \hline
\end{tabular}
\end{table}

\begin{table}[]
\scriptsize
\centering
\caption{Ablation Study on VOC2007 test set (\textbf{constrained})} 
\label{ablation2}
\begin{tabular}{lllllll}
\hline 
Random Location      &$\surd$&  &  &      &      &    \\
Center BBOX Location   &   &$\surd$&    &      &      &    \\
Gradient-based Location&  &  &  $\surd$&$\surd$& $\surd$& $\surd$     \\
Patch Selection and Refining    &             &      &  &$\surd$&     & $\surd$    \\
Stabilizing gradient  & &      &      & &$\surd$    & $\surd$    \\ \hline\\
Y $\overline{BS}$      &   0.874   &0.987& 0.989 & 0.996 & \textbf{0.999} & 0.997 \\
F $\overline{BS}$      &   0.604   &0.816& 0.950 & 0.983 & 0.950 & \textbf{0.984} \\
$\overline{AS}$        &   1.000  &1.000   & 1.000 & 1.689 & 1.000 & \textbf{1.692} \\
$\overline{OS}$        &   1.478  &1.803 & 1.939 & 3.348 & 1.949 & \textbf{3.351} \\ \hline
\end{tabular}
\end{table}

In this section, we demonstrate the effectiveness of our methods under both unconstrained and constrained conditions on VOC2007 test set(only using the images with the category of sheep) and report the results in Table \ref{ablation1} and \ref{ablation2}. For these two cases, we use two baselines: randomly selecting 2\% of the perturbed pixels and selecting the perturbations in the center of the instance boxes.

For the case without the constraint about the number of the connected perturbed regions, whether the initial locations of perturbations are selected, the final results are not much different. This also proves that key-pixels cannot be accurately found at the beginning of the attack process. On the contrary, our patch selection and refining scheme improves the performance and removes the inconsequential perturbations gradually which greatly improves the OS. Moreover, stabilizing gradient in an unconstrained condition also slightly improves the OS. Further, using both of them can hide 100\% objects from YOLO v4 and Faster R-CNN.

As for the case with the constraint about the number of the connected perturbed regions, the adversarial patches determined by gradient heat map achieve better performance, especially when attacking Faster R-CNN. Moreover, the proposed scheme and stabilizing gradient all improve BS and the former also removes a lot of inefficient perturbation. Combining them can hide 99.7\% objects from YOLO v4 and 98.4\% objects from Faster R-CNN within 10 patches. These results further demonstrate the effectiveness of the proposed method.

\section{Conclusion}

In this paper, we propose a novel refined patch-based attack method named RPAttack on general object detectors which can generate patches with much less perturbation, performs better than other works. In order to find the key-pixels for attack and remove inconsequential perturbation, we introduce a novel patch selection and refining scheme. To our knowledge, this is the first method to select key-pixels based on gradient adaptively. For a stable ensemble attack, we balance the gradients from detectors with different architectures to avoid over-optimizing one of them. Our RPAttack can achieve an amazing missed detection rate of 100\% for both Yolo v4 and Faster R-CNN, while only modifies 0.32\% pixels on VOC 2007 test set. The experimental results show the deep learning based detectors are extremely vulnerable to the adversarial patch attack, even if only very few pixels are modified. We hope our work can arouse more attention to the potential threats of the adversarial patch attack.

\bibliographystyle{IEEEbib}
\bibliography{icme2021template}

\end{document}